\documentclass{article}
\usepackage[final,nonatbib]{neurips_2018}

\usepackage[utf8]{inputenc} 
\usepackage[T1]{fontenc}    
\usepackage{hyperref}       
\usepackage{url}            
\usepackage{booktabs}       
\usepackage{amsfonts}       
\usepackage{nicefrac}       
\usepackage{microtype}      
\usepackage{amsmath,amssymb}
\usepackage{algorithmic}
\usepackage{subcaption}
\usepackage{graphicx}
\usepackage{tabularx} 

\usepackage{multirow}
\usepackage[boxruled,linesnumbered]{algorithm2e}

\DeclareMathOperator*{\argmin}{arg\,min}
\DeclareMathOperator*{\argmax}{arg\,max}

\newcommand{\Rnx}[2]{\MBB{R}^{#1\times #2}}

\newcommand{\K}{{\mathcal K}}
\newcommand{\MB}[1]{\mathbf{#1}}
\newcommand{\MBB}[1]{\mathbb{#1}}
\newcommand{\MBBR}[1]{\mathbb{R}^{#1}}
\newcommand{\MC}[1]{\mathcal{#1}}

\newcommand{\Tr}[0]{\mathrm{Tr}}

\newcommand{\I}{\MC{I}}

\newcommand{\T}{\top}
\newcommand{\Bsum}{\mathlarger\sum}

\newcommand{\smalleqb}[1]
{
	\begingroup
	\makeatletter
	\def
	\f@size{1}
	{#1}
    \endgroup
}  

\newtheorem{prop}{Proposition}
\newtheorem{theo}{Theorem}
\newtheorem{proof}{Proof}
\newtheorem{defin}{Definition}

\newcommand{\ncl}{p}
\newcommand{\nf}{f}
\newcommand{\ndic}{c}

\newcommand{\ny}{{\vec{x}}}
\newcommand{\nY}{\mathbf{X}}
\newcommand{\nxs}{\gamma}
\newcommand{\nx}{{\vec{\gamma}}}
\newcommand{\nX}{\mathbf{\Gamma}}

\newcommand{\nA}{\mathbf{U}}
\newcommand{\na}{{\vec{u}}}
\newcommand{\nas}{u}
\newcommand{\nT}{T_0}
\newcommand{\nh}{{\vec{l}}}
\newcommand{\nhs}{l}
\newcommand{\nH}{\mathbf{L}}

\newcommand{\nAlf}{\vec{\beta}}
\newcommand{\nalf}{{\beta}}
\newcommand{\phih}{\hat{\Phi}}

\newcommand{\method}{IMKPL~}

\title{Interpretable Multiple-Kernel Prototype Learning for\\ Discriminative Representation and Feature Selection}

\author{%
	Babak Hosseini
	\thanks{
		Preprint of the publication~\cite{hosseini2019intermulti}, as provided by the authors.
		The final publication is available at \url{https://dl.acm.org/proceedings.cfm}
	} \\
	Center of Cognitive Interactive Technology (CITEC)\\
	Bielefeld University, Germany\\
	\texttt{bhosseini@techfak.uni-bielefeld.de}	
	\And
	Barbara Hammer\\
	Center of Cognitive Interactive Technology (CITEC)\\
	Bielefeld University, Germany\\
	\texttt{bhammer@techfak.uni-bielefeld.de} \\
}

\begin{document}
\maketitle

\begin{abstract}
Prototype-based methods are of the particular interest for domain specialists and practitioners as they summarize a dataset by a small set of representatives.
Therefore, in a classification setting, interpretability of the prototypes is as significant as the prediction accuracy of the algorithm.
Nevertheless, the state-of-the-art methods make inefficient	trade-offs between these concerns by sacrificing one in favor of the other, 
especially if the given data has a kernel-based (or multiple-kernel) representation. 
In this paper, we propose a novel interpretable multiple-kernel prototype learning (IMKPL) to construct highly interpretable prototypes in the feature space, 
which are also efficient for the discriminative representation of 
the data.
Our method focuses on the local discrimination of the classes in the feature space and shaping the prototypes based on condensed class-homogeneous neighborhoods of data.
Besides, IMKPL learns a combined embedding in the feature space in which the above objectives are better fulfilled.
When the base kernels coincide with the data dimensions, 
this embedding results in a discriminative features selection.
We evaluate IMKPL on several benchmarks from different domains which demonstrate its superiority to the related state-of-the-art methods regarding both interpretability and discriminative representation.
\end{abstract}

\section{Introduction}	
Cognitive psychology claims that
human categorizes different data classes in his mind by finding their most representative prototypes (examples)~\cite{rosch1975cognitive}.
Inspired by that, prototype-based (PB) models have become popular machine learning methods used by domain specialists in many applications.
A supervised PB algorithm finds representatives in the input space,
and predicts the class label based on their distances to the given data point ~\cite{friedman2001elements}.
%
%
Apart from this simplicity, their decisions are highly explainable (e.g., for a practitioner) by the direct inspection of the prototypes to which each test data is assigned~\cite{hammer2014learning}.
The most popular prototype-based approaches are the self-organizing map (SOM)~\cite{Kohonen1995}, 
nearest prototype classifier~\cite{friedman2001elements}, 
learning vector quantization (LVQ)~\cite{Kohonen1995}, 
and prototype selection (PS)~\cite{bien2011prototype} which are all supervised methods except SOM.
%

Beside the discriminative performance of these methods,
another significant concern is to learn interpretable prototypes which can represent 
condensed data neighborhoods without any inter-class overlapping~\cite{friedman2001elements}.
Usually, this concern induces a trade-off between the discriminative and interpretative quality of the prototypes~\cite{bien2011prototype}, 
and more often, the model sacrifices one of them in favor of the other.
In addition, regardless of their reported efficiency and simplicity, 
they face difficulties when the classes have extensive overlapping 
or are distinct but linearly inseparable (e.g., XOR dataset).

In nowadays applications, it is common to observe non-Euclidean data settings, such as time-series and sequences. 
A practical solution is to compute  
a relational representation based on non-Euclidean pairwise dissimilarities between the data points~\cite{teng2017anomaly}.
Consequently, the kernel variants of prototype-based methods are designed by assuming a corresponding implicit mapping to
a reproducing kernel Hilbert space (RKHS).
In particular, kernel K-means~\cite{shawe2004kernel} and 
kernelized-generalized LVQ (KGLVQ)~\cite{schleif2011efficient} represent the well-known unsupervised and supervised PB algorithms respectively.
Nevertheless, kernel-based methods generally have weak interpretation as their prototypes are constructed by a broad inter-class set of data points~\cite{nova2014review}.
In contrast, some dissimilarity-based methods 
such as the PS algorithm select class-specific exemplars 
%
to increase the interpretability of the prototypes.
However, this choice restricts the discriminative power of the model. 
%
%
%

Applying different kernels on the inputs results in a multiple-kernel (MK) representation of the data which might carry non-redundant pieces of information about essential properties of the data~\cite{bach2004multiple,teng2017anomaly}.
Consequently, multiple-kernel learning approaches are designed to find an effective weighted combination of these base kernels that enhances the classification performance.
Besides, by choosing only the features with non-zero combination weights, one can perform a discriminative feature selection~\cite{wang2014feature}.
%
Nevertheless, the majority of MK methods focus on finding a non-realistic combined RKHS on which the data classes could be linearly (globally) separable~\cite{shawe2004kernel}.
To our knowledge, no MK method is designed particularly for prototype-based representations and specifically with a focus on the local separation of the classes.
%


Dictionary learning (DL) finds a set of dictionary atoms in the input space to reconstruct each data sample by a sparse weighted combination (sparse code) of them~\cite{rubinstein2008efficient}. 
The sparse representations can capture essential intrinsic characteristics of the dataset \cite{kim2010sparse} that are consistent between the training and testing distributions.
The supervised DL methods try to 
also preserve the label information in the sparse encoding ~\cite{guo2016discriminative,DBLP:conf/cikm/WuLZ18}.
Furthermore, some recent works similar to~\cite{wang2014feature,zhu2017multi} 
joined MK learning with DL in order to improve the reconstruction and discrimination quality of the dictionary by optimizing it on an efficient combined RKHS.
Although one can consider the dictionary atoms as a set of representative prototypes,
no multiple-kernel (or single-kernel) dictionary learning algorithm (MKDL)
have that explicit focus in its design.
%
Hence, 
their learned dictionary atoms either suffer from the weak 
interpretation or cannot discriminatively represent the classes.
\textbf{Contributions}:
In this paper, we propose interpretable multiple-kernel prototype learning (IMKPL) algorithm
to obtain a discriminative prototype-based model
for multiple-kernel data representations.
%
%
We construct our framework upon the basic model of
multiple-kernel dictionary learning, such that 
each data sample can be represented by a set of prototypes.
%
%
More specifically, our work contributes to the current state-of-the-art in the following aspects:
\begin{itemize}	
	\item We extend the application of prototype-based learning to MK data representations.
	\item Our model effectively learns interpretable prototypes based on the class-specific local neighborhoods they represent.
	\item Our prototype-learning framework focuses on local discrimination of the classes on the combined RKHS to mitigate their global overlapping.
\end{itemize}

In the next section, we discuss the relevant work. Our proposed IMKPL algorithm is introduced in Section \ref{sec_sc_cl}, and we explain its optimization scheme in Section~\ref{sec:optim}. The empirical evaluations are presented in Section~\ref{sec_exp}, and we provide the conclusion of work in the final section.

\section{Preliminaries}\label{sec:rel}
We define the data matrix $\nY=\left[\ny_1,...,\ny_N\right] \in \mathbb{R}^{d\times N}$ containing $N$ training samples and 
$\nH=[\nh_1,\dots,\nh_N] \in\mathbb{R}^{p \times N}$
as the corresponding label matrix in a $\ncl$-class setting.
Each $\nh_i$ is a zero vector with $\nhs_{qi}=1$ if $\ny_i$ is in class $q$. 
As conventions, 
$\vec v_i$ denotes the $i$-th column in a given matrix $\MB{V}$, 
$v_{ji}$ and $v_{j}$ refer to the $j$-th entries in vectors $\vec v_i$ and $\vec v$ respectively,
and $\vec{v}^j$ denotes the $j$-th row in $\MB{V}$.

Assuming $\nf$ implicit non-linear mappings $\{\Phi_i(\ny)\}_{i=1}^\nf$ projecting $\nY$ onto $\nf$ individual RKHSs \cite{bach2004multiple}, it is possible to compute the weighted kernel function $\phih(\ny)$ based on the following scaling
\begin{equation}
\begin{array}{l}
\phih(\ny)=
[\sqrt{\nalf_1} \Phi_1^\top(\ny),\cdots,\sqrt{\nalf_{\nf}} \Phi_{\nf}^\top(\ny)]^\top,
\end{array}
\label{eq:mk}
\end{equation}
where $\nAlf$ is the combination vector which results in a combined RKHS corresponding to $\phih(\ny)$. 
By assuming that each $\Phi_i(\ny)$ is related to a base kernel function $\K_i(\ny_t,\ny_s)=\Phi_i^\top(\ny_t)\Phi_i(\ny_s)~\forall t,s$, 
we derive the weighted kernel $\hat{\K}(\ny_t,\ny_s) ~\forall t,s$ as
\begin{equation}
\hat{\K}(\ny_t,\ny_s)=\sum_{i=1}^{\nf} \nalf_i \K_i(\ny_t,\ny_s)
=\phih(\ny_t)^\T \phih(\ny_s).
\label{eq:K_alf}
\end{equation}
By choosing $\nAlf \in \MBB{R}_{\geq 0}$,
one can interpret the 
entries of $\nAlf$
as the relative importance of each base kernel in the obtained MK representation $\phih(\ny)$~\cite{gonen2011multiple}.
For ease of reading, we denote
the Gram matrix $\hat{\K}(\nY,\nY)$ and the vector $\hat{\K}(\ny_i,\nY)$ by $\hat{\K}$ and $\hat{\K}(i,:)$ respectively.

The goal of multiple-kernel dictionary learning (MKDL) is to 
find an optimal MK dictionary $\phih(\MB{D})$ on the combined RKHS to reconstruct the inputs as $\phih(\nY)\approx \phih(\MB{D}) \nX$ in this space.
The columns of $\nX=[\nx_1,\dots,\nx_N] \in \mathbb{R}^{\ndic\times N}$ contain the corresponding sparse codes 
which are desired to have sparse non-zeros entries~\cite{Aharon2006}.
A basic MKDL framework can be formulated as a variant of the following  
%
\begin{equation}
\begin{array}{ll}
\underset{\nX,\nA}{\min}&
\| \phih(\nY)-\phih(\nY)\nA\nX\|_F^2\\
\mathrm{s.t.} & \|\nAlf\|_1=1, ~\nalf_i \in
\mathbb{R}_{\geq 0},~\|\nx_i\|_0 \le \nT,
\end{array}
\label{eq:mklsrc}
\end{equation}
where the objective term
$\MC{J}_{rec}= \| \phih(\nY)-\phih(\nY)\nA\nX\|_F^2$
measures the reconstruction quality of the data on the RKHS.
%
In Eq.~(\ref{eq:mklsrc}),
$\|.\|_0$ and $\|.\|_F^2$ denote the cardinality and $F$-norm respectively,
and the dictionary is modeled as
$\phih(\MB{D})=\phih(\nY)\nA$ where $\nA \in \mathbb{R}^{N\times \ndic}$ is the dictionary matrix~\cite{VanNguyen2013}. 
Hence, each column of $\nA$ defines a linear combination of data points in the feature space while it can be optimized in the input space.
The constraint $\|\nAlf\|_1=1$ 
applies an affine combination of the base kernels and also prevents   
the trivial solution $\nAlf=0$.
%
%
%
%
The role of $\nAlf$ in $\phih(\nY)$ is to enhance the discriminative power of the learned dictionary atoms $\{\phih(\nY)\na_i\}_{i=1}^\ndic$ by
increasing the dissimilarity between the different-label columns in $\phih(\nY)$.

Although $\MC{J}_{rec}$ is a common term 
in MKDL methods,
it varies 
based on the formulation of MK or DL part.
In \cite{thiagarajan2014multiple}, the vector $\nAlf$ was individually optimized to improve the linear separability of the classes on the RKHS.
%
In contrast, \cite{shrivastava2015multiple} jointly optimized $\{\nA,\nAlf\}$
by pre-defining class-isolated sub-dictionaries in $\nA$ and 
enforcing the orthogonality of 
each class to the dictionaries of other classes on the RKHS;
and \cite{zhu2017multi} utilized an analysis-synthesis 
class-isolated dictionary model
along with a low-rank constraint on $\nX$.


Compared to 
these frameworks, 
we explicitly shape
the dictionary atoms as interpretable prototypes, 
to improve local representation and discrimination of the classes effectively.
%
However, none of the major MKDL methods adequately provide such PB model.

\section{Interpretable Multiple-Kernel Prototype Learning}\label{sec_sc_cl} 
We want to learn an MK dictionary that its constituent prototypes (atoms) reconstruct the data while presenting discriminative, interpretable characteristics regarding the class labels.
Explicitly, we aim for the following specific objectives:
\newline
\textbf{O1:}
Assigning prototypes to the local neighborhoods in the classes to efficiently discriminate them on the RKHS regarding their class labels (Figure~\ref{fig:lmkl}-d).
\newline
\textbf{O2:}
Learning prototypes which can be interpreted by the condensed class-specific neighborhoods they represent (Figure~\ref{fig:lp}-b)
\newline
\textbf{O3:}
Obtaining an efficient MK representation of the data to assist the above objectives and to improve the local separation of the classes on the resulted RKHS (Figure~\ref{fig:lmkl}). 
\begin{defin}
	Each $\ny$ is represented by the set  of prototypes $\{\phih(\nY)\na_i\}_{i \in I}$ on the combined RKHS
	if $\|\phih(\ny)-\phih(\nY)\nA\nx\|_2^2<\epsilon$ for a small $\epsilon>0$ and $\forall i \in  I,~\nxs_i\neq0$. 
	\label{def:pres}
\end{defin}
Based on Definition~\ref{def:pres}, we call $\{\na_i\}_{i=1}^\ndic$ the prototype vectors to represent the columns of $\phih(\nY)$,
%
and we propose the interpretable multiple-kernel prototype learning algorithm
to learn them while adequately
addressing the above objectives. 
IMKPL has the novel optimization scheme of:
\begin{equation}
\begin{array}{ll}
\underset{\nAlf,\nX,\nA}{\min}
&\|\phih(\nY)-\phih(\nY)\nA\nX\|_F^2
+\lambda \MC{J}_{dis}
+\mu \MC{J}_{ls}+\tau \MC{J}_{ip}\\
\mathrm{s.t.} & \|\nx_i\|_0 < \nT,
~~\|\nAlf\|_1=1,
~~\|\phih(\nY)\na_i\|_2^2=1,\\
& \|\na_i\|_0 \le \nT,~~ \nas_{ji}, \nalf_i, \nxs_{ji} \in \mathbb{R}_{\geq 0},\\
\end{array}
\label{eq:dksrc}
\end{equation}
in which $\lambda$, $\tau$, and $\mu$ are trade-off weights. 
The cardinality and non-negativity constraints on $\{\nA,\nX\}$
coincide with the model structure $\phih(\nY)\nA$ \cite{hosseini2018conf}.
They motivate each prototype $\phih(\nY)\na_i$ to be formed by sparse contributions from similar training samples in $\phih(\nY)$ to increase their interpretability~\cite{bien2011prototype}.
Although each $\na_i$ is loosely shaped from the local neighborhoods in the RKHS, it cannot fulfill the objectives \textbf{O1} and \textbf{O2} on its own (Figure~\ref{fig:lp}-a).
Also, having $\|\phih(\nY)\na_i\|_2^2=1$ prevents the solution of $\na_i$ from being degenerated \cite{rubinstein2008efficient}.

In the following subsections,  
while addressing the objectives \textbf{O1}-\textbf{O3},
we explain the novel terms $\{\MC{J}_{dis}, \MC{J}_{ls}, \MC{J}_{ip}\}$. 

\subsection{Discriminative Loss $\MC{J}_{dis}(\nA,\nX,\nAlf)$} \label{sec:dsrc}
\begin{figure}
		\centering		
		\includegraphics[width=.91\linewidth]{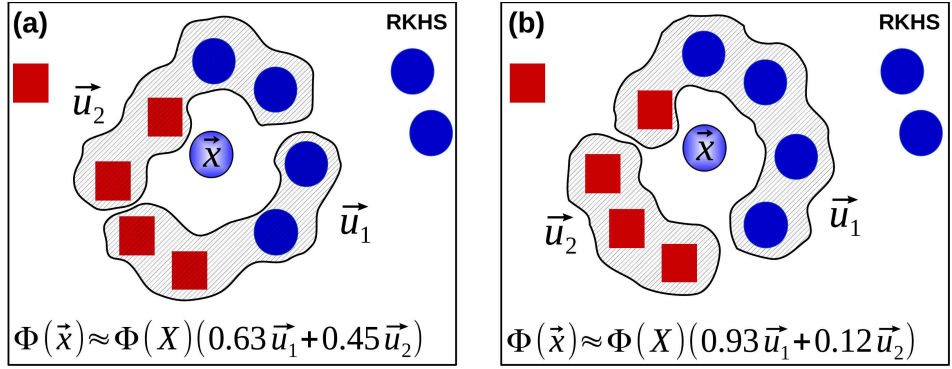}
	\caption
	{ 
		The effect of $\MC{J}_{dis}$ in Eq.~(\ref{eq:dksrc}).
		(a): When $\lambda=0$, prototypes $(\na_1,\na_2)$ (hatched selections) are shaped and reconstruct $\phih(\ny)$ by its neighboring samples 
		from both of the classes (circles and squares).
		(b): When $\lambda\neq0$, these prototypes are formed s.t. $\phih(\ny)$ is approximately represented by $\na_1$ which is mostly shaped from its local, same-class neighbors (circles).}
	\label{fig:lp}
\end{figure}

By rewriting $\phih(\nY)\nA\nx=\phih(\nY)\vec{h}$, the vector
$\vec{h}\in \MBBR{N}$ reconstructs $\phih(\ny)$ based on other samples in $\phih(\nY)$.
%
Hence, by aiming for \textbf{O1}, we learn the prototype vectors $\{\na_i\}_{i=1}^\ndic$ such that they represent each $\phih(\ny)$ with a corresponding vector $\vec{h}$ 
using mostly
the local same-class neighbors of $\phih(\ny)$.
Accordingly, we define the loss term $\MC{J}_{dis}$ as:
\begin{equation}
\begin{array}{l}
\MC{J}_{dis}(\nA,\nX,\nAlf)=\\
\frac{1}{2}
\overset{N}{\underset{i=1}{\Bsum}}
[\overset{N}{\underset{s=1}{\sum}}
\na^s\nx_i 
(\nh_i^\T \nh_s\|\phih(\ny_i)-\phih(\ny_s) \|_2^2

+ \|\nh_i-\nh_s \|_2^2)].
\end{array}
\label{eq:o_dis}
\end{equation}
%
\begin{prop}
	The objective $\MC{J}_{dis}$ in Eq.~(\ref{eq:o_dis}) has its minimum if 
	$\forall \ny_i$, $\phih(\ny_i)\approx \phih(\nY)\nA\nx_i$ 
	s.t.
	$\forall t:\nxs_{ti} \neq 0, \forall s:\nas_{st}\neq 0$,
	$\nh_i=\nh_s\text{ and } \|\phih(\ny_i)-\phih(\ny_s)\|_2^2  
	\approx 0$.
	%
	\begin{proof}
		The objective term $\MC{J}_{dis}$ is constructed upon summation and multiplication of non-negative elements. Hence, its global minima would lie where $\MC{J}_{dis}(\nA,\nX)=0$ holds.
		This condition can be fulfilled if for each $\nx_i$:
		$$
		[\overset{N}{\underset{s=1}{\sum}}
		\na^s 
		(\nh_i^\T \nh_s\|\phih(\ny_i)-\phih(\ny_s) \|_2^2
		+ \|\nh_i-\nh_s \|_2^2)]\nx_i=0.
		$$
		Since the trivial solution $\nx_i=0$ is avoided due to $\MC{J}_{rec}$ in Eq.~(\ref{eq:dksrc}), we can find a set $\MC{I}$ s.t. $\forall t\in \MC{I}, \nxs_{ti}\neq 0$ holds.
		Therefore, $\forall t\in \MC{I}, \sum_{s=1}^N\nas_{st}\Omega_{si}=0$, where
		$$
		\Omega_{si}=\nh_i^\T \nh_s\|\phih(\ny_i)-\phih(\ny_s) \|_2^2
		+ \|\nh_i-\nh_s \|_2^2.
		$$
		It is clear that 
		$$\Omega_{si}=
		\begin{cases} 2 & \nh_i \neq \nh_s \\
		\|\phih(\ny_i)-\phih(\ny_s) \|_2^2 &  \nh_i = \nh_s,
		\end{cases}
		$$
		which means that $\forall s$, $\nas_{st}\Omega_{si}=0$	holds in either of the following cases: 
		\begin{enumerate}
			\item $\nas_{st}=0$, meaning that the data point $\ny_s$ does not contribute to the $t$-th prototype (e.g., consider the squares in Figure~\ref{fig:lp}-b which are not a part of $\na_1$) .
			\item $\na_{t}$ uses $\ny_s$ that lies in the same class as $\ny_i$ (e.g., the circles in Figure~\ref{fig:lp}-b as the main constituents of $\na_1$).
		\end{enumerate}
		Putting all the above conditions together, $\MC{J}_{dis}=0$ happens only if in case of the condition described by the proposition. 						
	\end{proof}
	\label{prop:dis}
\end{prop}
Although Proposition \ref{prop:dis} describes the ideal situations, in practice, it is common to observe $\|\phih(\ny_i)-\phih(\ny_s) \|_2^2 <\epsilon$ for a small, non-negative $\epsilon$ when $\ny_s$ is among the neighboring points of $\ny_i$.
This condition results in small non-zero minima for $\MC{J}_{dis}$.
Besides, for a given $\ny_i$, if its cross-class neighbors lie closer to its same-class neighbors, $\Omega_{si}$ obtains higher values by choosing $\ny_s$ s.t. $\ny_s \neq \nh_i$ in favor of better minimizing $\MC{J}_{rec}$ (e.g., the squares in Figure~\ref{fig:lp}-b which is a part of $\na_1$).

Based on Proposition~\ref{prop:dis}, minimizing $\MC{J}_{dis}$ enforces the framework in Eq.~(\ref{eq:dksrc}) to learn $\nA$ 
such that each prototype $\phih(\nY)\na_i$ is shaped by a concentrated neighborhood 
in RKHS which can provide a discriminative representation for its nearby samples.
%
However, $\MC{J}_{dis}$ is still flexible to tolerate small cross-class contributions in the representation of each $\ny_i$ 
in case of overlapping between the classes.
For example, in Figure~\ref{fig:lp}-b, 
although a square sample has contributed to the reconstruction of $\ny$ 
(due to their small distance), 
$\ny$ is still represented mostly by samples of its own class (\textit{circles}). 
%
\subsection{Interpretability Loss $\MC{J}_{ip}(\nA)$}
\begin{defin}
	Prototype $\phih(\nY)\na_i$ is interpretable as a local representative of the class $q$
	if $\forall t:\nas_{ti}\neq 0$, $\ny_t$ belongs to a concentrated neighborhood in the RKHS and $\frac{\nh^q\na_i}{\|\nH \na_i\|_1} \approx 1$.
%
	\label{def:ip}
\end{defin}
%
%
%
When the class-overlapping is subtle,
minimizing $\MC{J}_{dis}$ can result in interpretable prototypes 
(e.g., in Figure~\ref{fig:lp}-b, $\na_1$ can still be interpreted as a local representative for the \textit{circle} class).
However, a considerable overlapping of the classes results in having more than one large entries in each $\vec{s}=\nH \na_i$ (similar to $\na_1$ in Figure~\ref{fig:lp}-a). 
%
Therefore, 
to better satisfy objective \textbf{O2},
we define 
$
	\MC{J}_{ip}(\nA)=\|\nH\nA\|_1,
$	
such that its minimization reduces $\|\vec{s}\|_1$ for each prototype vector.
So, this term 
together with 
$\MC{J}_{dis}$
results in a significantly sparse $\nH \na_i$,
such that 
${\nh^q\na_i}/{\|\nH \na_i\|_1}$ 
obtains a value close to 1,
which improves the interpretability of each $\phih(\nY)\na_i$ according to Definition~\ref{def:ip}.
\subsection{Local-Separation Loss $\MC{J}_{ls}(\nAlf)$} \label{sec:LMK}
\begin{figure}[!t]
		\centering		
		\includegraphics[width=0.91\linewidth]{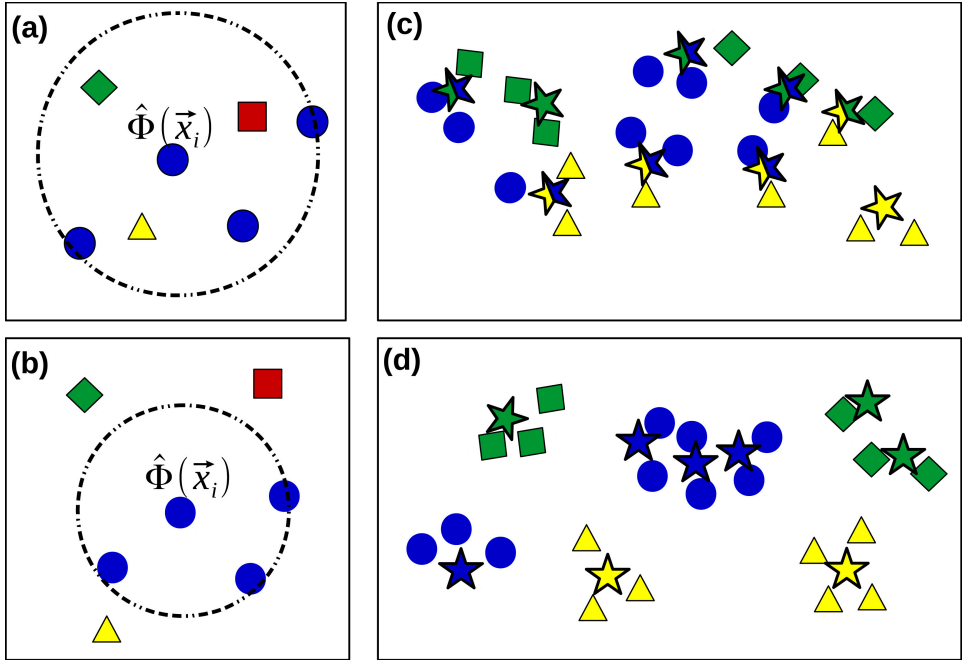}			
	\caption{
		Effect of $\MC{J}_{ls}$ on local separation of each $\phih(\ny_i)$ 
		from its different-label neighbors in RKHS when $k=4$ (\textbf{b} compared to \textbf{a}),
		which concentrates the classes locally (\textbf{d} compared to \textbf{c}) and improves
		the interpretation of the prototypes $\{\phih(\nY)\na_i\}_{i=1}^\ndic$ (the stars) by the class-neighborhood to which they are assigned (their colors).}
	\label{fig:lmkl}
\end{figure}

According to Eqs.~(\ref{eq:mk}) and (\ref{eq:dksrc}),
the weighting vector $\nAlf$ is already incorporated into $\MC{J}_{rec}$ and $\MC{J}_{dis}$ via its role in $\phih(\nY)$. 
Hence, minimizing them w.r.t. to $\nAlf$ optimizes the combined embedding in the features spaces to fulfill the objectives \textbf{O1} and \textbf{O2} better.
Besides, as an effective complement,
we optimize $\nAlf$ to separate the classes locally in $k$-size neighborhoods. 
%
%
We propose 
$\MC{J}_{ls}$ as the following novel, convex loss:
\begin{equation}	
	\begin{array}{l}
		\MC{J}_{ls}(\nAlf)=
		\underset{i=1}{\sum^N}
		\big[\underset{s\in \MC{N}_i^k}{\sum} \|\phih(\ny_i)-\phih(\ny_s)\|_2^2 		
		+ \underset{s\in \overline{\MC{N}_i^k}}{\sum} \phih(\ny_i)^\T\phih(\ny_s) \big],		
	\end{array}	
	\label{eq:const}
\end{equation}
where $\MC{N}_i^k$ 
specifies the same-label $k$-nearest neighbors of $\ny_i$ on the RKHS, 
and $\overline{\MC{N}_i^k}$ is its corresponding $k$-size set for 
the different-label neighbors of $\ny_i$.
Eq.~(\ref{eq:const}) reaches its minima
when for each $\ny_i$: 1. The summation of its distances to the nearby same-label points is minimized, and 2. It is dissimilar from the nearby data of other classes (Figure~\ref{fig:lmkl}-b). 
Therefore, having $\MC{J}_{ls}$ 
in conjunction with other terms in Eq.~(\ref{eq:dksrc}) 
makes the classes locally condensed and distinct from each other,
which facilitates  
learning better interpretable, discriminative prototypes (Figure~\ref{fig:lmkl}-d).
%
%
In the next section, we explain how to solve the optimization problem of Eq.~(\ref{eq:dksrc}) efficiently.
\section{Optimization Scheme of IMKPL}\label{sec:optim}
After re-writing the optimization problem of Eq.~\ref{eq:dksrc} using the given definitions 
for $\{\MC{J}_{dis}, \MC{J}_{ls}, \MC{J}_{ip}\}$,
we optimize its parameters $\{\nA,\nX,\nAlf\}$ 
by adopting the alternating optimization scheme.
\begin{prop}
	Denoting $\nA \in \Rnx{N}{\ndic}$, $\nX \in \Rnx{\ndic}{N}$, $\nAlf\in\MBB{R}^\nf$,	
	and $g(\nA,\nX,\nAlf)=
	\|\phih(\nY)-\phih(\nY)\nA\nX\|_F^2\\
	+\lambda \frac{1}{2}
	\overset{N}{\underset{i=1}{\Bsum}}
	[\overset{N}{\underset{s=1}{\sum}}
	\na^s\nx_i 
	(\nh_i^\T \nh_s\|\phih(\ny_i)-\phih(\ny_s) \|_2^2	
	+ \|\nh_i-\nh_s \|_2^2)]\\
	+\mu \underset{i=1}{\sum^N}
	\big[\underset{s\in \MC{N}_i^k}{\sum} \|\phih(\ny_i)-\phih(\ny_s)\|_2^2 		
	+ \underset{s\in \overline{\MC{N}_i^k}}{\sum} \phih(\ny_i)^\T\phih(\ny_s)
	+\tau \|\nH\nA\|_1$,
    the objective function $g(\nA,\nX,\nAlf)$ is multi-convex in terms of $\{\nX,\nA,\nAlf\}$.
    \begin{proof}
    	Sketch:
    	Each of the defined functions in $\{\MC{J}_{rec},\MC{J}_{dis}, \MC{J}_{ls}, \MC{J}_{ip}\}$ is convex w.r.t any individual member of
    	$\{\nA,\nX,\nAlf\}$ due to having a positive semi-definite assumption for $\K_i ~~\forall i=1,\dots,\nf$ or the specific definition of 
    	the given objective function ($\MC{J}_{ls}$ is linear in terms of $\nAlf$, and $\MC{J}_{rec}$ is the F-norm function).
   	\end{proof}	 
   \label{prop:convx}
\end{prop}
Benefiting from Proposition~\ref{prop:convx},
at each of the following alternating steps, we update only one of the parameters while fixing the others (Algorithm~\ref{alg:mksrc}). 
The derivation of the following sub-problems is provided in the supplementary material.
\subsection{Updating the Matrix of Sparse Codes $\nX$}\label{sec:opt_X}
By fixing $\{\nA,\nAlf\}$, using Eq.~(\ref{eq:K_alf}), and removing the constant terms, we reformulate Eq.~(\ref{eq:dksrc}) w.r.t. each $\nx_i$ as:
\begin{equation}
\begin{array}{ll}
\underset{\nx_i}{\min} 
& {\nx_i}^\top (\nA^\top \hat{\K}\nA)\nx_i
+[\lambda\tilde{\K}(i,:) -2\hat{\K}(i,:)]\nA\nx_i\\
\mathrm{s.t.} & ~~ \|\nx_i\|_0 < \nT, ~~ \nxs_{ji} \in \MBB{R}_{\geq 0},
\end{array}
\label{eq:optx2}
\end{equation}  
where $\tilde{\K}=\MB{1}-(\nH^\T\nH)\odot\hat{\K}$ while $"\odot"$ denotes the Hadamard product operator.
This optimization problem is a non-negative quadratic programming problem with a cardinality constraint on $\nx_i$. 
The matrix $\nA^\top \hat{\K}\nA$ is positive semidefinite (PSD) because $\hat{\K}$ is PSD and $\nA$ is non-negative. Hence, Eq.~(\ref{eq:optx2}) is a convex problem, and we efficiently solve it by 
proposing the Non-negative Quadratic Pursuit (NQP) algorithm (Algorithm.~\ref{alg:NQP}).
%
Hence, we update the columns of $\nX$ individually.
\subsection{Updating Prototype Matrix $\nA$}
Similar to the approximation of $\nX$, the prototype vectors  $\na_i$ are updated sequentially. 
We rewrite the reconstruction objective $\MC{J}_{rec}$ in Eq.~(\ref{eq:dksrc}) as
\begin{equation}
\| \hat{\Phi}(\nY) \mathbf{E}_i-\hat{\Phi}(\nY)\na_i \nx^i\|_F^2, \quad \MB{E}_i=(\mathbf{I}-{\sum}_{j\neq i} \na_j \nx^j),
\label{eq:ei}
\end{equation}
where $\MB{I}\in \MBB{R}^{N \times N}$ is an identity matrix.
By using Eq.~(\ref{eq:ei}) and 
writing $\MC{J}_{dis}$ in terms of $\na_i$,
we reformulate Eq.~(\ref{eq:dksrc}) as
\begin{equation}
\resizebox{.8891\linewidth}{!}{$
\begin{array}{ll}
\underset{\na_i}{\min} 
& {\na_i}^\top (\nx^i \nx^{i^\top} \hat{\K})\na_i
+[\nx^i(-2\MB{E}_i^\T \hat{\K}+\lambda\tilde{\K})
+\tau \vec{1}^\T \nH]\na_i\\
\mathrm{s.t.} & ~~ \|\na_i\|_0 < \nT, ~~\|\phih(\nY)\na_i\|_2^2=1,~~ \nas_{ji} \in \mathbb{R}_{\geq 0}.
\end{array}
$}
\label{eq:opta}
\end{equation}
%
Analogous to Eq.~(\ref{eq:optx2}), this is a convex non-negative quadratic problem in terms of $\na_i$ with a hard limit on $\|\na_i\|_0$.
Hence, we update the prototype vectors $\{\na_i\}_{i=1}^\ndic$ by solving Eq.~(\ref{eq:opta}) using the NQP algorithm.
After updating each $\na_i$, we normalize it as 
$\na_i=\frac{\na_i}{\|\phih(\nY)\na_i\|_2}=\frac{\na_i}{\sqrt{\na_i^\top\hat{\K}\na_i}}.$
\subsection{Updating Kernel Weights $\nAlf$}
By normalizing each base kernel $\K_i$ in advance, 
we can simplify Eq.~(\ref{eq:dksrc}) to the following linear programming (LP) problem
\begin{equation}
\begin{array}{ll}
\underset{\nAlf}{\min}
& (\vec{\MC{E}}_{rec}+\lambda\vec{\MC{E}}_{dis}+\mu \vec{\MC{E}}_{ls})^\T \nAlf\\
\mathrm{s.t.} & 
\sum_{i=1}^{\nf}{\nalf_i}=1,~ \nalf_i \in \mathbb{R}_{\geq 0},
\end{array}
\label{eq:alf}
\end{equation}
where we derive the entries of $\vec{\MC{E}}_{rec}$, $\vec{\MC{E}}_{dis}$, and $\vec{\MC{E}}_{ls}$ by incorporating Eq.~(\ref{eq:K_alf}) into the terms $\MC{J}_{rec}$, $\MC{J}_{dis}$, and $\MC{J}_{ls}$ respectively.
We compute their $l$-th entries ($l=1,\dots,\nf$) as
\begin{equation}
\resizebox{.89\linewidth}{!}{$
\begin{array}{l}
{\MC{E}}_{rec}(l)=\Tr[\K_l(\MB{I}-2\nA\nX)+\nX^\top \nA^\top \K_l \nA \nX],\\
{\MC{E}}_{dis}(l)=\Tr(\tilde{\K}_l\nA\nX),\\
{\MC{E}}_{ls}(l)=\Bsum_{i=1}^N
\underset{s\in \MC{N}_i^k}{\sum}
[2-2\K_l(\ny_i,\ny_s) ]
+ \underset{s\in \overline{\MC{N}_i^k}}{\sum}
\K_l(\ny_i,\ny_s),
\end{array}
$}
\label{eq:alfa_obj}
\end{equation}
where $\tilde{\K}_l$ is derived by computing $\tilde{\K}$ 
while replacing $\K$ with $\K_l$, and $\Tr(.)$ denotes the trace operator.
Therefore, we can efficiently solve the LP in Eq.~(\ref{eq:alf}) using linear solvers 
~\cite{strayer2012linear}.
Algorithm~\ref{alg:mksrc} provides an overview of all the optimization steps for our IMKPL framework.

\begin{algorithm}[t] 
	\caption{{\small Interpretable  Multiple-Kernel Prototype Learning}} 
	\label{alg:mksrc} 
\begin{flushleft}
	\textbf{Parameters}: Weights $\{\lambda,\mu,\tau\}$, sparsity $\nT$, neighborhood size $k$\\
	\textbf{Input}: Label matrix $\nH$, kernel functions $\{\K_i(\nY,\nY)\}_{i=1}^\nf$\\
	\textbf{Output}: Prototype vectors $\{\na_i\}_{i=1}^\ndic$, kernel weights $\nAlf$,
	encodings $\nX$\\
	\textbf{Initialization}: Computing $\{\tilde{\K},\{\tilde{\K}_i\}_{i=1}^\nf,\vec{\MC{E}}_{ls}\}$, $\nAlf=\vec{\MB{1}}$\\
\end{flushleft}	
	\begin{algorithmic}[1] 
	\REPEAT 
	\STATE{Computing $\hat{\K}(\nY,\nY)=\sum_{i=1}^{\nf}\nalf_i \K_i(\nY,\nY)$\\		
	\STATE Updating $\nX$ based on Eq.~(\ref{eq:optx2}) using NQP\\
	\STATE Updating $\nA$ based on Eq.~(\ref{eq:opta}) using NQP\\
	\STATE Updating $\nAlf$ based on Eq.~(\ref{eq:alf}) using an LP solver
	}
    \UNTIL{convergence}
	\end{algorithmic}
\end{algorithm}	
\subsection{Representation of the Test Data}
To represent (reconstruct) a test data $\ny_{test}$ by the trained $\nA$ and $\nAlf$,
we compute the sparse code $\nx_{test}$ using Eq.~(\ref{eq:optx2}) while setting $\lambda=0$. The relational values of the entries in $\nx_{test}$ show the main prototypes which are used to represent $\ny_{test}$.
%
\subsection{Non-negative Quadratic Pursuit}\label{sec:NQP}
Consider a quadratic function $f(\nx):=\frac{1}{2}\nx^\top \MB{Q} \nx+\vec{c}^\top \nx$, in which
$\nx \in \MBB{R}^n$, $\vec{c} \in \MBB{R}^{n}$, and $\MB{Q} \in \MBB{R}^{n \times n}$ is a Hermitian positive semidefinite matrix.
Non-negative quadratic pursuit algorithm (NQP) is an extended form of the Matching Pursuit problem \cite{Aharon2006} and is inspired from by \cite{lee2006efficient}. 
Its objective is to minimize $f(\nx)$ approximately in an NP-hard optimization problem similar to
\begin{equation}
\begin{array}{ll}
\nx=&\underset{\nx}{\argmin} \frac{1}{2} \nx^\top \MB{Q} \nx+\vec{c}^\top \nx\\
&s.t. ~~~~ \|\nx\|_0 \leq \nT ~,~  \nxs_i \ge 0~~\forall i 
\end{array}
\label{eq:qp_x}
\end{equation}
where at most $\nT \ll n$ elements from $\nx$ are permitted to be positive while all other elements are forced to be zero.

As presented in Algorithm \ref{alg:NQP},
at each iteration of NQP we compute $\nabla_\nx f(\nx)$ to guess about the next promising dimension of $\nx$ (denoted as $\nxs_j$) which may lead to the biggest decrease in the current value of $f(\nx_{\I})$;
where $\I$ denotes the set of currently chosen dimensions of $\nx$ based on the previous iterations.
We look for $\nx\ge0$ solutions, and also the current value of $\nx$ entries for new dimensions are zero; therefore, similar to the Gauss-Southwell rule in coordinate descent optimization~\cite{nesterov2012efficiency} we choose the dimension $j$ which is related to the smallest negative entry of $\nabla_\nx f(\nx)$ as
\begin{equation}
j=\underset{j \in S}{\argmin} ~{\vec{q}}^{\top}_{j}\nx+c_j \qquad s.t. ~
{\vec{q}}^{\top}_{j}\nx+c_j < 0
\label{eq:j}
\end{equation}
where $\vec{q}_j$ is the $j$-th column of $\MB{Q}$.
Then by adding $j$ to $\I$, the resulting unconstrained quadratic problem is solved using the closed form solution $\nx_{\I}={-\MB{Q}_{\I\I}}^{-1}\vec{c}_{\I}$, and generally, we repeat this process until reaching $\|\nx\|_0=\nT$ criterion. 
Notation $\MB{Q}_{\I\I}$ and $\vec{c}_{\I}$ denote the principal submatrix of $\MB{Q}$ and the subvector of $\vec{c}$ respectively corresponding to the set $\I$.

To preserve non-negativity of the solution $\nx$ in each iteration $t$ of NQP, 
in case of having a negative entry in $\nx_{\I}^t$, 
a simple line search is performed between $\nx_{\I}^t$ and $\nx_{\I}^{(t-1)}$.
The line search chooses the nearest zero-crossing point to $\nx_{\I}^{(t-1)}$ on the connecting line between $\nx_{\I}^{(t-1)}$ and $\nx_{\I}^{t}$.

In addition, to reduce the computational cost, we use the Cholesky factorization $\MB{Q}_{\I\I}=\MB{L}\MB{L}^\top$ \cite{van1996matrix} to compute $\nx$ with a back-substitution process.


Furthermore, because matrix $\MB{Q}$ in equations \eqref{eq:qp_x} is PSD, its principal sub-matrix $\MB{Q}_{\I\I}$ should be either PD or PSD theoretically \cite{johnson1981eigenvalue}, where the first case is a requirement for the Cholesky factorization.   
However, by choosing $\nT << rank(Q)$ in practice, we have never confronted a singular condition.
Nevertheless, to avoid such rare conditions, we do a non-singularity check for the selected dimension $j$ which is to have $q_{jj}\neq v^\top v$ right after obtaining $v$ (1st Cholesky step in Algorithm \ref{alg:NQP}). In case the resulted $v$ does not fulfill that condition, we choose another $j$ based on (\ref{eq:j})


\begin{algorithm} 
	\caption{Non-negative Quadratic Pursuit} 
	\label{alg:NQP} 
	
	\begin{algorithmic} 
		\STATE {\bfseries Parameters:} $\nT$, $\epsilon:$ stopping threshold.
		\STATE {\bfseries Input:} $\MB{Q} \in \MBB{R}^{n \times n},c\in \MBB{R}^{n}$ when $f(\nx)=\frac{1}{2} \nx^\top \MB{Q} \nx+\vec{c}^\top \nx$.
		\STATE {\bfseries output:} An approximate solution $\nx$.
		\STATE {\bfseries Initialization:} $\nx=0$ , $\MC{I} =\{\}$ , $\MC{S} =\{1,...,n\}$ , $t=1$.
		\REPEAT	
		
		\STATE 	$j=\underset{j \in S}{\argmin} ~{\vec{q}}^{\top}_{j}\nx+c_j$ \qquad $s.t. ~$
		${\vec{q}}^{\top}_{j}\nx+c_j < 0$\\
		\STATE 	\textbf{if} $j = \varnothing$ \textbf{then} Convergence.\\
		\STATE 	$\MC{I}:=\I \cap j$;\\
		\STATE 	\small $\vec{q}_{\I j}:=$ created via selecting rows $\I$ and column $j$ of matrix $\MB{Q}$.\\
		\STATE 	\small $\vec{c}_{\I}:=$ a subvector of $c$ based on selecting entries $\I$ of vector $\vec{c}$.\\
		\IF{$t>1$}
		\STATE			$v:= \text{Solve for } v ~\big\{\MB{L}v=\vec{q}_{\I j} \big\}$;\\
		\STATE			$\MB{L}:=
		\begin{bmatrix}
		\MB{L} & 0 \\
		v^\top & \sqrt{q_{jj}-v^\top v}
		\end{bmatrix}
		$\\
		\ELSE
		\STATE		{$\MB{L}=q_{jj}$}
		\ENDIF
		\STATE		$\nx_{\I}^{t}:= \text{Solve for } x~ \big\{\MB{L}\MB{L}^\top x=\vec{c}_{\I} \big\}$;\\
		\IF {$\exists{j\in \MBB{N}};~ (\nxs_j^t<0)$}
		\STATE	$\nx_{\I}^t:=$ the nearest zero-crossing to $\nx_{\I}^{(t-1)}$ via a line search.\\
		\STATE	$\MC{S}:=\MC{S}-\{\text{zeros entries in } \nx_{\I}^{t}\}$
		\ENDIF
		\STATE	$\MC{S}:=\MC{S}-j$\\
		\STATE		$t=t+1$
		\UNTIL {$(\MC{S} = \{\}) ~\vee ~(\|\nx\|_0=\nT) ~\vee ~ (\frac{1}{2} \nx^\T Q \nx +c^\T \nx <\epsilon)$}
		\STATE	Convergence.	
	\end{algorithmic}
\end{algorithm}	

\subsubsection{The Convergence of NQP}
NQP does not guarantee the global optimum as it is a greedy selection of rows/columns of matrix $\MB{Q}$ to provide a sparse approximation of the NP-hard problem in (\ref{eq:qp_x}); nevertheless, its convergence to a local optimum point is guaranteed. 
\begin{theo}
	The Non-negative Quadratic Pursuit algorithm (Algorithm~\ref{alg:NQP}) converges to a local minimum of Eq.~(\ref{eq:qp_x}) in a limited number of iterations.
\begin{proof}
The algorithm consists of 3 main parts: 
\begin{enumerate}
	\item Gradient-based dimension selection
	\item Closed form solution
	\item Non-negative line search and updating $\MC{I}$. 
\end{enumerate}
It is clear that the closed-form solution $\nx$ via selecting a negative direction of the gradient $\nabla_\nx f(\nx)$ always reduces the current value of $f(\nx^t)$ as $\nx^t$ has to be non-negative and initially $\nxs_j=0$. 
In addition, The zero-crossing line search in iteration $t$ can guarantee to strictly reduce the value of $f(\nx^{(t-1)})$. 
It finds a non-negative $\nx^t_{new}$ between the line connecting $\nx_{\I}^{(t-1)}$ to $\nx_{\I}^{t}$, and since $f(\nx)$ is convex, $f(\nx_{new}^t)<f(\nx_{\I}^{(t-1)})$ 

Consequently, each of the steps guarantees a monotonic decrease in the value of $f(\nx)$, therefore if $\|\nx^{(t+i)}\|_0 > \|\nx^{(t)}\|_0 \implies f(\nx^{(t+i)}) < f(\nx^{(t)})$. 
Also, the algorithm structure guarantees that in any iteration $t$, $\MC{I}_t \neq \MC{I}_i ~\forall i <t $ meaning that NQP never gets trapped into a loop of repeated dimension selections. Furthermore, we have $\|\nx\|_0 \leq \	nT$, meaning that the total number of possible selections in $\MC{I}$ is bounded. Concluding from the above, the NQP algorithm converges in a limited number of iterations.
\end{proof}
\label{theo:nqp}
\end{theo}
\subsubsection{The Computational Complexity of NQP}\label{sec:nqp_comx}
We can calculate the computational complexity of NQP by considering its individual steps. 
Iteration $t$ contains computing $\MB{Q}\nx + \vec{c}$ ($nt+t$ operation), finding minimum of $\nabla_\nx f(\nx)$ w.r.t the negative constraint ($2n$ operations), computing $v$ ($t^2$ operation for the $t\times t$ back-substitution), computing $\nx_{\I}^t$ (two back-substitutions resulting in $2t^2$ operation), and checking negativity of entries of $\nx_{\I}^t$ along with the probable line-search which has $3t$ operations in total.
Hence, the total runtime of each iteration is bounded by 
$$\MB{T}_{iter}=(n+4)\nT+2n+\nT^2.$$

Although the algorithm looks for maximum $\nT$ elements to estimate $\nx$, 
due to the non-negativity constraint the algorithm converges in a small number of iterations
(usually below 20) which is independent of the size of $\MB{Q}$ or $\nT$.
Also by considering that in practice $\nT \ll n$, the algorithm's computational complexity is $\MC{O}(n\nT)$, 
which is smaller than its comparable algorithm, the Orthogonal Matching Pursuit (as $\MC{O}(n\nT^2+\nT^3)$)\cite{Aharon2006}.
\subsection{Complexity and Convergence of IMKPL}
%
In order to calculate the computational complexity of IMKPL per iteration, we analyze the update of each $\{\nX,\nA,\nAlf\}$ individually.
In each iteration, the update of $\nX$ and $\nA$ are done using the NQP algorithm, which has the time complexity of $\MC{O}(n\nT)$, where $n$ is the number of dimensions in the quadratic problem.
Also, we set $\ndic=\ncl\nT$ as an effective choice in our model, and in practice the maximum number of non-zero elements of $\nx^i$
in Eq.~(\ref{eq:opta}) is smaller than $\frac{N}{\ncl}$.

Therefore, optimizing $\nX$ and $\nA$ leads to 
$\MC{O}(\ncl N\nT^2+\ncl \nT N^2)$ and 
$\MC{O}(\ncl N\nT^2+\nT N^2+\ncl N)$ computational costs respectively,
and optimizing $\nAlf$ with an LP solver has the computational complexity of 
$\MC{O}(2t\nf+\nf N^2+\nf kN)$, where $t$ is the convergence iteration of the LP-solver. 
The time-consuming matrix multiplications of Eq.~(\ref{eq:alfa_obj}) are already carried out while solving 
Eqs.~(\ref{eq:optx2}),(\ref{eq:opta}).

%
%

As in the implementations we observe/choose $\ncl,\nT,k<<N$ (eps. for large-scale datasets), 
the computational complexity of IMKPL in each iteration is 
approximately $\MC{O}(\nf N^2+N^2)$.
Therefore, \method is more scalable than its alternative MK algorithms 
~\cite{zhu2017multi,thiagarajan2014multiple,shrivastava2015multiple} which have complexity close to $\MC{O}(\nf N^3)$.
%
%

We provide the following proof regarding the convergence of Algorithm~\ref{alg:mksrc}: 
\begin{theo}
	The iterative updating procedure in Algorithm~\ref{alg:mksrc} converges to a locally optimal point in a limited number of iterations.	
	\begin{proof}
		Based on Proposition~\ref{prop:convx} and Theorem~\ref{theo:nqp},  
		each optimization sub-problem in Algorithm~\ref{alg:mksrc} reduces the objective function of Eq.~\ref{eq:dksrc} monotonically.
		In addition, all the individual objective terms in Eq.~\ref{eq:dksrc} are bounded from below by zero according to their definitions. Therefore, convergence to at least a local minimum solution is guaranteed under a limited number of iterations.  		 
		\end{proof}
\end{theo}
We present the converge curve of the IMKPL algorithm in the experiments (Sec.~\ref{sec:conv}) showing its convergence in less than 20 iterations for all the selected real-datasets.
The implementation code of NQP optimization algorithm is available online \footnote{https://github.com/bab-git/NQP}.	 

\section{Experiments}\label{sec_exp}
%
%
%
%
%
%
%
%
For our experiments, we implement IMKPL on 
%
the vectorial datasets 
$\{$\texttt{CLL\_SUB\_111}, 
\texttt{TOX\_171},
\verb|Isolet|$\}$\footnote{http://featureselection.asu.edu/datasets.php},
and \verb|rcv1|(subset1-topic classification)\footnote{http://mulan.sourceforge.net/datasets-mlc.html};
and on multivariate time-series (MTS) datasets
$\{$\texttt{PEMS}, \texttt{AUSLAN}$\}$\footnote{https://archive.ics.uci.edu/ml/index.php},
and the \texttt{Utkiect} Human action dataset~\cite{xia2012view}.
Table~\ref{tab:data} provides the specific characteristics of these datasets.
The code of IMKPL algorithm and supplementary material is available online
\footnote{https://github.com/bab-git/IMKPL}.
\begin{table}
	\centering
	\caption{Information and settings of selected datasets.}	
	\begin{tabular}{lcccccccc  } %
		\toprule	
		Dataset & $\nT$ & $\lambda$ & $\mu$ & $\tau$& size & feature & class\\
		\midrule
		 \texttt{CLL\_SUB}&15 &0.2&0.1&0.2&111&11340&3\\
		
		 \texttt{TOX\_171}&18 &0.1&0.3&0.1&171&5748&4\\
		
		\verb|Isolet|  &20 &0.2&0.2&0.3&1560&617&26\\
		
		\verb|UTKinect| &5&0.1&0.1&0.2&200&60&10\\
				
		\verb|PEMS| &20&0.2&0.3&0.1&440&963&7\\
		
		\verb|AUSLAN| &12&0.3&0.1&0.2&2500&128&95\\
		
		\verb|rcv1|(subset1) &15&0.1&0.2&0.3&6000&944&101\\
		\bottomrule		
	\end{tabular} 			
	\label{tab:data} 
\end{table}
\subsection{Experimental Setup}
To compute the base kernels for vectorial datasets, we use the Gaussian kernel 
$\K_i(\ny_s,\ny_t)=exp(-\mathcal{D}(\ny_s^i,\ny_t^i)^2/\delta_i)$
where $\ny^i$ denotes the $i$-th dimension of $\ny$, and 
$\mathcal{D}(\ny^i_s,\ny^i_t)=\|\ny^i_s-\ny^i_t\|_2$ as the Euclidean distance between each pair of $\{\ny_s,\ny_t\}$ in that dimension.
The scalar $\delta_i$ is equal to the average of $\mathcal{D}(\ny_s^i,\ny_t^i)^2$ over all data points.
For \verb|rcv1|(subset1) text classification dataset, we obtain the vectorial representation based on the word-embedding used by \cite{pivovarova2018comparison}.
For MTS datasets, we compute each $\K_i$ via employing the global alignment kernel (GAK)~\cite{cuturi2007kernel} for the $i$-th dimension of the time-series. 
Exceptionally for \verb|Utkiect|, prior to the application of GAK, we use the pre-processing from ~\cite{vemulapalli2014human} to obtain the Lie Group representation.

We compare our proposed method to the following state-of-the-art prototype-based learning or multiple-kernel dictionary learning methods:
AKGLVQ~\cite{schleif2011efficient},
PS~\cite{bien2011prototype},
MKLDPL~\cite{zhu2017multi}, 
DKMLD~\cite{thiagarajan2014multiple},
and MIDL~\cite{shrivastava2015multiple}.
The AKGLVQ algorithm is the sparse variant of the kernelized-generalized LVQ~\cite{schleif2011efficient}, and for the PS algorithm, we use its distance-based implementation.
These two algorithms are implemented on the average-kernel inputs ($\nAlf=\vec{1}$).
Hence, we also implement ISKPL as the single-kernel variant of IMKPL on that input representation.
\\
\textbf{Note:}
We elusively select the baselines
which can be evaluated according to our 
specific research objectives (\textbf{O1-O3}).
%
%
%
%
%

We perform 5-fold cross-validation on the training set to tune the hyper-parameters $\{\lambda,\mu,\nT,\tau\}$ in Eq.~(\ref{eq:dksrc})
which are reported in Table~\ref{tab:data}.
%
We carry out a similar procedure regarding the parameter tuning of other baselines. 
For IMKPL,
we determine the number of prototypes as $\ndic=\ncl\nT$ and the neighborhood radius $k=\nT$.
As the rationale, the constraint $\|\na_i\|_0\le \nT$ and the term $\MC{J}_{dis}$ in Eq.~(\ref{eq:dksrc}) make each $\na_i$ effective mostly on its $\nT$-radius neighborhood. 
%
In practice, choosing $\lambda=\mu=\tau \in [0.2~0.4]$ is a good working setting for IMKPL to initiate the parameter tuning (e.g., Figure~\ref{fig:sens}).
\subsection{Evaluation Measures}
%
To evaluate the quality of the learned prototypes on the resulted RKHS (based on $\{\nA,\nAlf\}$), we utilize the following measures which coincide with our objectives \textbf{O1-O3}.
%
%
\subsubsection{Interpretability of the Prototypes ($IP$)}
As discussed in Section \ref{sec_sc_cl}, we have two main preferences regarding the interpretability of each prototype $\phih(\nY)\na_i$: 
1.~Its formation based on class-homogeneous data samples.
2.~Its connection to local neighborhoods in the feature space.
Therefore we use the following $IP$ term to evaluate the above criteria based on the values of the prototype vectors $\{\na_i\}_{i=1}^\ndic$:
%
$$
\begin{array}{l}
IP=100 \times \frac{1}{\ndic}
\overset{\ndic}{\underset{i=1}{\sum}}
\frac{\nh^q\na_i}{\|\nH \na_i\|_1}
exp({-\underset{s,t}{\sum}
	 \nas_{si}\nas_{ti}\|\phih(\ny_s)-\phih(\ny_t)\|_2^2}),
\end{array}
$$
in which $q=\underset{q}{\argmax}~\nh^q\na_i$
is the class to which the $i$-th prototype is assigned.
The first part of this equation obtains the maximum value of $1$ if each $\na_i$
has its non-zero entries related to only one class of data, while the exponential term becomes $1$ (maximum) if those entries correspond to a condensed neighborhood of points in RKHS.
Hence, $IP$ becomes close to $100 \%$ if both of the above concerns are sufficiently fulfilled.
For the PS algorithm, we measure $IP$ based on the samples inside $\epsilon$-radius of each prototype~\cite{bien2011prototype}.
%
\subsubsection{Discriminative Representation ($DR$)}
In order to properly evaluate how discriminative each prototype $\phih(\nY)\na_i$ is we define the discriminative representation term as
$
DR=100 \times \frac{1}{\ndic}
\sum_{i=1}^\ndic
\frac
{\sum_{s:\nh_s=q} \nxs_{is}}{\|\nx^i\|_1},
$
where $q$ is the same as in $IP$ measure, and $\nX$ is computed based on the test set. 
Hence, $DR$ becomes $100\%$ (maximum) if each prototype $i$
which is assigned to class $q$ 
only represents (reconstructs) that class of data;
in other words, the prototypes provide exclusive representation of the classes.
%
The $DR$ measure does not fit the models of AKGLVQ and PS algorithms.
\subsubsection{Classification Accuracy of Test Data ($Acc$)}
For each test data $\ny_{test}$,
we predict its class as
$
q=\underset{q}{\argmax}~~ \nh^q \nA \nx_{test},
$
meaning that the $q$-th class provides the most contributions in the reconstruction of $\ny_{test}$.
%
Accordingly, we compute the average of accuracy value $Acc=100\frac{\# \text{correct predictions}}{N}$
%
over 10 randomized test/train selections for each dataset.	
\begin{figure}[b!]
	\centering		
	\includegraphics[width=1\linewidth,height=10cm]{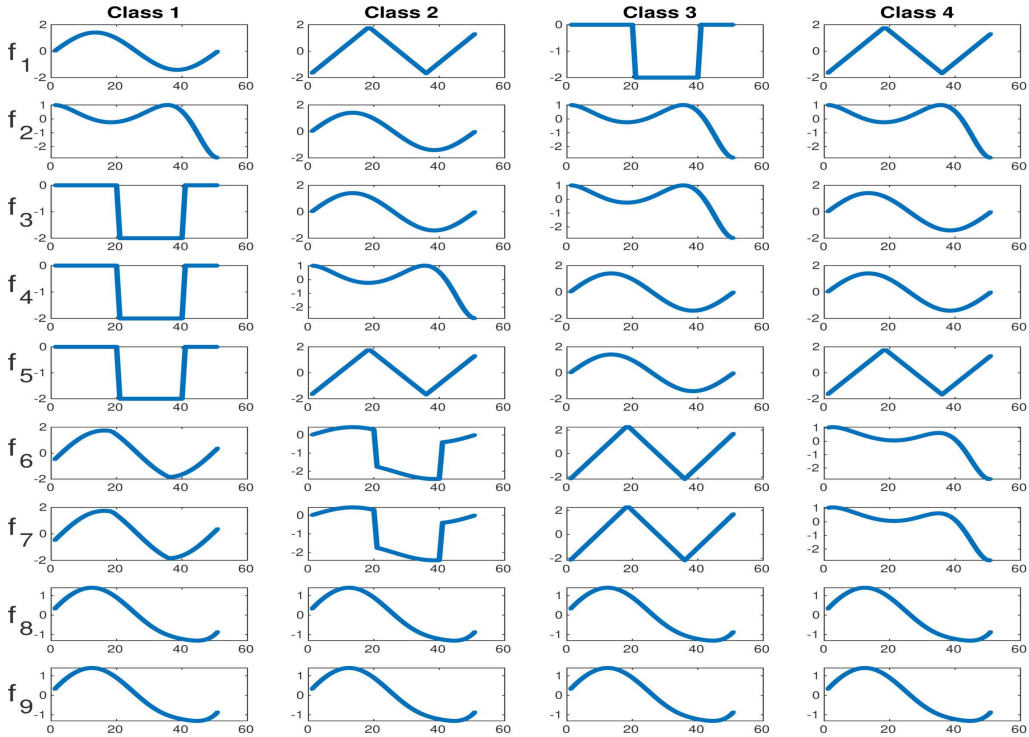}	
	\caption{Four classes of synthetic multi-variant time-series. Column $i$: A time-series exemplar from the $i$-th class. Row $i$: The $i$-th feature of the time-series.}
	\label{fig:synt_all_0}
\end{figure}

\subsection{Results: Synthetic Dataset}
In order to illustrate the feature selection performance of the IMKPL algorithm, a 4-classes dataset of multi-variant time-series is designed using variations of simple 1-dimensional curves.
As depicted in Fig.\ref{fig:synt_all_0}, the first 5 features (rows) in the $i$-th data exemplar ($i$-th column) follow a specific pattern related to class $i$. However, for each class, $f_7$ is the replicate of $f_6$ with slight variations, and the last two features ($f_8,f_9$) are identical through all the samples.

As a result, 9 individual kernel functions $\{\K_i\}_{i=1}^9$ are computed corresponding to features $\{f_i\}_{i=1}^9$. Despite the simplicity of the dataset for the classification task, we are interested in studying the performance of IMKPL regarding final feature weightings.

After application of IMKPL, the data is classified with $100 \%$ accuracy and the following $\nAlf$, which selects only the features $\{f_3,f_4,f_6\}$: 
$$\nAlf= [0,~0,~0.68,~0.50,~0,~0.54,~0,~0,~0]^\top$$
As a result, the last two identical features $\{f_8,f_9\}$ were ruled out as they were totally irrelevant to the discriminative LMK objective. 
However, these two features could be ideal choices to have a small reconstruction term $\mathcal{J}_{rec}$ in Eq.~(4).
Similarly, the weight of $f_7$ is 0 in $\nAlf$ as its kernel function is similar to $\K_6$.
However, $\K_6$ along with 2 other features $\{f_3,f_4\}$ have remained as they are determined to be useful for the discriminative representation. Features $\{f_1,f_2\}$ are removed due to the redundancy of $\{\K_1,\K_2\}$ comparing to others in providing a locally-separable representation. 
This experiment demonstrates how IMKPL can decide on the importance of the features based on their role in having a discriminative representation.

\subsection{Results: Efficiency of the Prototypes}
\begin{table}[b]
	\caption{Comparison of baselines regarding $IP(\%)$ and $DR(\%)$.}
	\centering
		\footnotesize
	\begin{tabularx}{0.7\textwidth}{lcc|cc|cc|cc} %
		\toprule				
		Methods & $IP$&  $DR$ & $IP$&  $DR$ & $IP$&  $DR$& $IP$&  $DR$ \\		
		\midrule
		&\multicolumn{2}{c|} {\footnotesize\texttt{CLL\_SUB}} 
		&\multicolumn{2}{c|} {\footnotesize \texttt{TOX\_171} } 
		&\multicolumn{2}{c|} {\footnotesize\verb|Isolet| }
		&\multicolumn{2}{c} {\footnotesize\verb|rcv1|(subset1)}\\				
		\midrule
		\textbf{IMKPL \scriptsize(ours)} 	&\textbf{91}&\textbf{75}&\textbf{95}&\textbf{89}&\textbf{96}&\textbf{90}& \textbf{94}&\textbf{82}\\
		\textbf{ISKPL \scriptsize(ours)} 	&88&70&93&79&94&81&88&79\\		
		MKLDPL 				   			&75&57&82&66&79&63&71&67\\
		DKMLD 				   			&67&51&71&60&72&57&65&64\\
		MIDL 				   			&66&50&69&60&69&54&61&57\\
		AKGLVQ							&69&--&76&--&78&--&68&--\\
		PS								&78&--&80&--&84&--&78&--\\
		\toprule
		&\multicolumn{2}{c|} {\footnotesize\texttt{UTKinect}} 
		&\multicolumn{2}{c|} {\footnotesize \texttt{PEMS} } 
		&\multicolumn{2}{c|} {\footnotesize\verb|AUSLAN| }\\		
		\midrule
		\textbf{IMKPL \scriptsize(ours)} 	&\textbf{96}&\textbf{91}&\textbf{95}&\textbf{88}&\textbf{97}&\textbf{87}\\		
		\textbf{ISKPL \scriptsize(ours)} 	&92&82&91&80&95&79\\		
		MKLDPL 				   			&78&60&77&59&80&70\\
		DKMLD 				   			&74&52&70&56&73&59\\
		MIDL 				   			&69&50&67&58&71&61\\
		AKGLVQ							&77&--&72&--&75&--\\
		PS								&79&--&76&--&81&--\\
		\bottomrule		
	\end{tabularx} 				
	\\{\scriptsize The best result (\textbf{bold}) is according to a two-valued t-test at a $5\%$ significance level}.
	\label{tab:dr} 
\end{table}
In Table~\ref{tab:dr}, we compare the baselines regarding 
the interpretability and discriminative qualities of their trained prototypes.
Considering the $IP$ values, IMKPL significantly outperforms both the MKDL and prototype-based learning algorithms.
For the \verb|rcv1| dataset, our method has a margin of $23\%$ compared to the best baseline algorithm (MKLDPL).
Also, the ISKPL algorithm obtains higher 
interpretability performances than the single-kernel and multiple-kernel baselines, which shows the effectiveness of the prototype leaning parts of the design ($\MC{J}_{dis}$ and $\MC{J}_{ip}$).
Besides, the difference between the $IP$ values of ISKPL and IMKPL signifies the role of the $\MC{J}_{ls}$ objective in enhancing the interpretation of IMKPL's prototypes by learning a suitable MK representation.
Other algorithms show weak results in learning class-specific and locally concentrated prototypes.

We observe similar behaviors by comparing the algorithms based on the $DR$ measure.
Table~\ref{tab:dr} shows that the prototypes learned by IMKPL are more efficient regarding the exclusive representation of the classes on a combined RKHS.
For instance, IMKPL outperforms MKLDPL (best baseline) with the $DR$ margin of $31\%$ on \verb|UTKinect| dataset.
\subsection{Results: Discriminative Feature Selection}
%
%
\begin{table}[t]
	\caption{Comparison of baselines regarding $Acc~(\%)$ and $\|\nAlf\|_0$.}	
	\centering
		\footnotesize
	\begin{tabularx}{0.79\textwidth}{lcc|cc|cc|cc} %
		\toprule				
		\footnotesize Methods & $Acc$&  $\|\nAlf\|_0$ & $Acc$&  $\|\nAlf\|_0$ & $Acc$&  $\|\nAlf\|_0$& $Acc$&  $\|\nAlf\|_0$ \\		
		\midrule
		&\multicolumn{2}{c|} {\footnotesize\texttt{CLL\_SUB}} 
		&\multicolumn{2}{c|} {\footnotesize \texttt{TOX\_171} } 
		&\multicolumn{2}{c|} {\footnotesize\verb|Isolet| }
		&\multicolumn{2}{c} {\footnotesize\verb|rcv1|(sub1)}\\
		\midrule
		\textbf{IMKPL\tiny(ours)} 	&\textbf{81.73}&204&\textbf{97.21}&\textbf{72}&\textbf{97.75}&141&\textbf{87.46}&220\\
		\textbf{ISKPL\tiny(ours)} 	&77.95&--&88.07&--&90.11&--&81.91&--\\		
		MKLDPL 				   			&79.39&310&94,72&347&95,21&224&83.32&384\\
		DKMLD 				   			&78.13&\textbf{101}&90,49&230&92,74&\textbf{126}&82.51&\textbf{180}\\
		MIDL 				   			&77,24&452&87,63&571&91,66&236&82.02&510\\	
		AKGLVQ							&74.66&--&86.21&--&88.32&--&81.57&--\\
		PS								&74.03&--&82.47&--&86.36&--&79.43&--\\
		\toprule
		&\multicolumn{2}{c|} {\footnotesize\texttt{UTKinect}} 
		&\multicolumn{2}{c|} {\footnotesize \texttt{PEMS} } 
		&\multicolumn{2}{c} {\footnotesize\verb|AUSLAN| }\\		
		\midrule
		\textbf{IMKPL\tiny(ours)} 	&\textbf{98.82}&\textbf{14}&\textbf{95.32}&89&\textbf{95.81}&\textbf{21}\\
		\textbf{ISKPL\tiny(ours)} 	&90.32&--&88.65&--&88.92&--\\
		MKLDPL 				   			&94.32&32&91,11&65&92.39&40\\		
		DKMLD 				   			&91.64&30&90,23&\textbf{62}&89.21&36\\
		MIDL 				   			&91,01&47&87,51&158&88.45&79\\
		AKGLVQ							&88.75&--&85.13&--&85.35&--\\
		PS								&85.89&--&84.54&--&83.65&--\\
		\bottomrule		
	\end{tabularx} 				
	\\{\scriptsize The best result (\textbf{bold}) is according to a two-valued t-test at a $5\%$ significance level}.
	\label{tab:fs} 
\end{table}
Each base kernel $\K_i$ is derived from one dimension of the
data. Therefore, we evaluate the discriminative feature selection performance
of the algorithms 
by comparing $\|\nAlf\|_0$ and $Acc$ among them.
%
%
As presented in Table~\ref{tab:fs}, IMKPL has the best prediction accuracies on all datasets. It outperforms other baselines with relatively significant $Acc$-margins (e.g., $4.21\%$ compared to MKLDPL on \verb|PEMS|).

Regarding $\|\nAlf\|_0$, IMKPL obtains the smallest set of selected features on three of the datasets. 
It particularly shows a significant feature selection performance on 
\texttt{TOX\_171} by obtaining $97.21\%$ accuracy while selecting 72 features out of the total 5748 dimensions.
Regarding other datasets (e.g., \texttt{CLL\_SUB} and \verb|Isolet|),
considering the $Acc$ values too reveals that 
the multiple-kernel optimization of IMKPL (role of $\nAlf$ in Eq.~(\ref{eq:dksrc})) 
finds an efficient set of features that lead to 
a discriminative PB model with a high performance (but not necessarily the smallest set).
This role is further noticeable when we compare IMKPL to ISKPL.
%

On the other hand, comparing the prediction accuracy of ISKPL to AKGLVQ and PS 
(as the major prototype-based learning methods)
demonstrates the significant discriminative performance of our algorithm in this domain. 
Besides, even though ISKPL obtained lower $Acc$ values than MKLDPL and DKMLD (as it does not optimize $\nAlf$), its higher $DR$ values show 
the effectiveness of its design ($\MC{J}_{dis}$ and $\MC{J}_{ip}$) 
regarding our expectations
from a prototype-based representation.
%
%
%
%
\subsection{Detail Analysis of Prototypes}
It is a common feature for many prototype-based methods to fix the number of prototypes for each class of data through the training phase
(e.g., MKLDPL, DKMLD, and AKGLVQ). However, as a common observation in real-world datasets, data classes are not distributed homogeneously.
Even having the same number of data per class, their local distributions can be significantly diverse. 

\begin{figure}[t]
	\centering		
	\includegraphics[width=.81\linewidth]{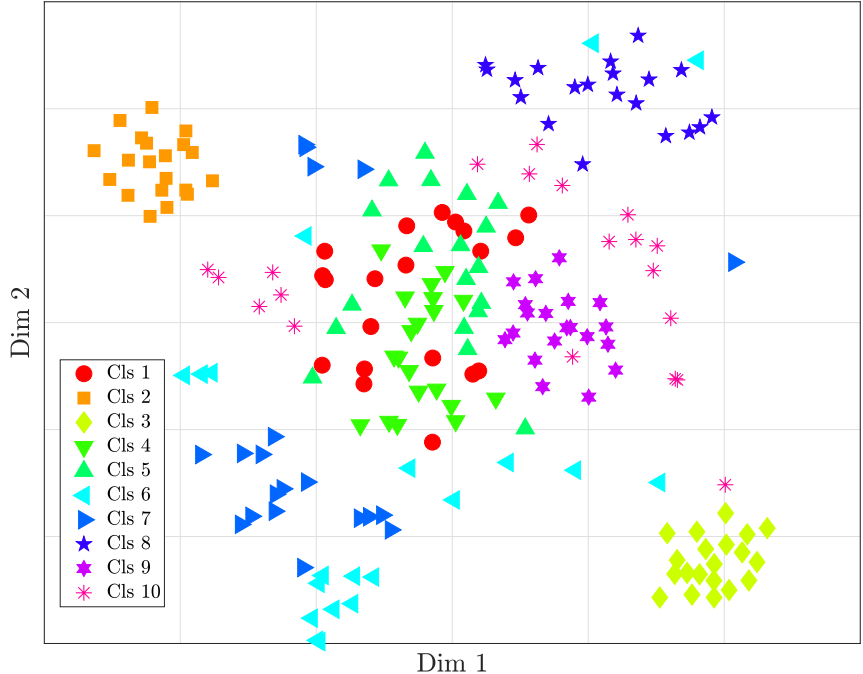}
	\caption
	{2-dimensional embedding of the \texttt{UTKinect} dataset (based on the average-kernel) which visualizes the relative overlapping of the classes (colored figure).}
	\label{fig:freq}
		\vspace{-0.2cm}
\end{figure}

In our \method model, 
although we decide in advance about the total number of prototypes to learn for each dataset as $\ndic=\ncl\nT$,
IMKPL automatically assigns the proper number of prototypes to each class of data to fulfill the defined objectives \textbf{O1}-\textbf{O3} better.
As reported in Table~\ref{tab:freq}, we examined the frequency of learned prototypes per class on the \verb|UTKinect| dataset, 
which shows a notable variation among them. 
Also, by considering the 2-dimensional embedding of the \verb|UTKinect| dataset (using the t-SNE algorithm~\cite{Maaten2008})
in Figure~\ref{fig:freq},
it is clear that IMKPL assigns more prototypes to classes which suffer from significant overlapping (e.g., \textit{pick up} and \textit{carry})
and fewer representatives to the more condensed classes (e.g., \textit{sit down} and \textit{stand up}).
\begin{table}[b]
	\centering
	\caption{Number of prototypes assigned to each class of the \texttt{UTKinect} dataset.}	
	\footnotesize
	\begin{tabularx}{0.7\textwidth}{Xcccccc} %
		\toprule	
		Classes &1&2&3&4&5&\\
		\midrule
		Names   &{ walk}& { sit down}&{ stand up}&{ pick up}&{ carry}& \\
		Prototypes &7	&2	&2	&8	&8	&\\		
		\toprule	
		Classes &6&7&8&9&10&All\\
		\midrule
		Names   &{ throw}&{ push}&{ pull}&{ wave}&{ clap}&\\
		Prototypes 	&6&5	&4	&3	&5&50\\		
		\bottomrule				
	\end{tabularx} 				
	\label{tab:freq} 
\end{table}

\subsubsection{Visualization of the Learned Kernel}
\begin{figure}[t]
	\begin{center}
		\minipage{0.45\textwidth}
		\includegraphics[width=\linewidth]{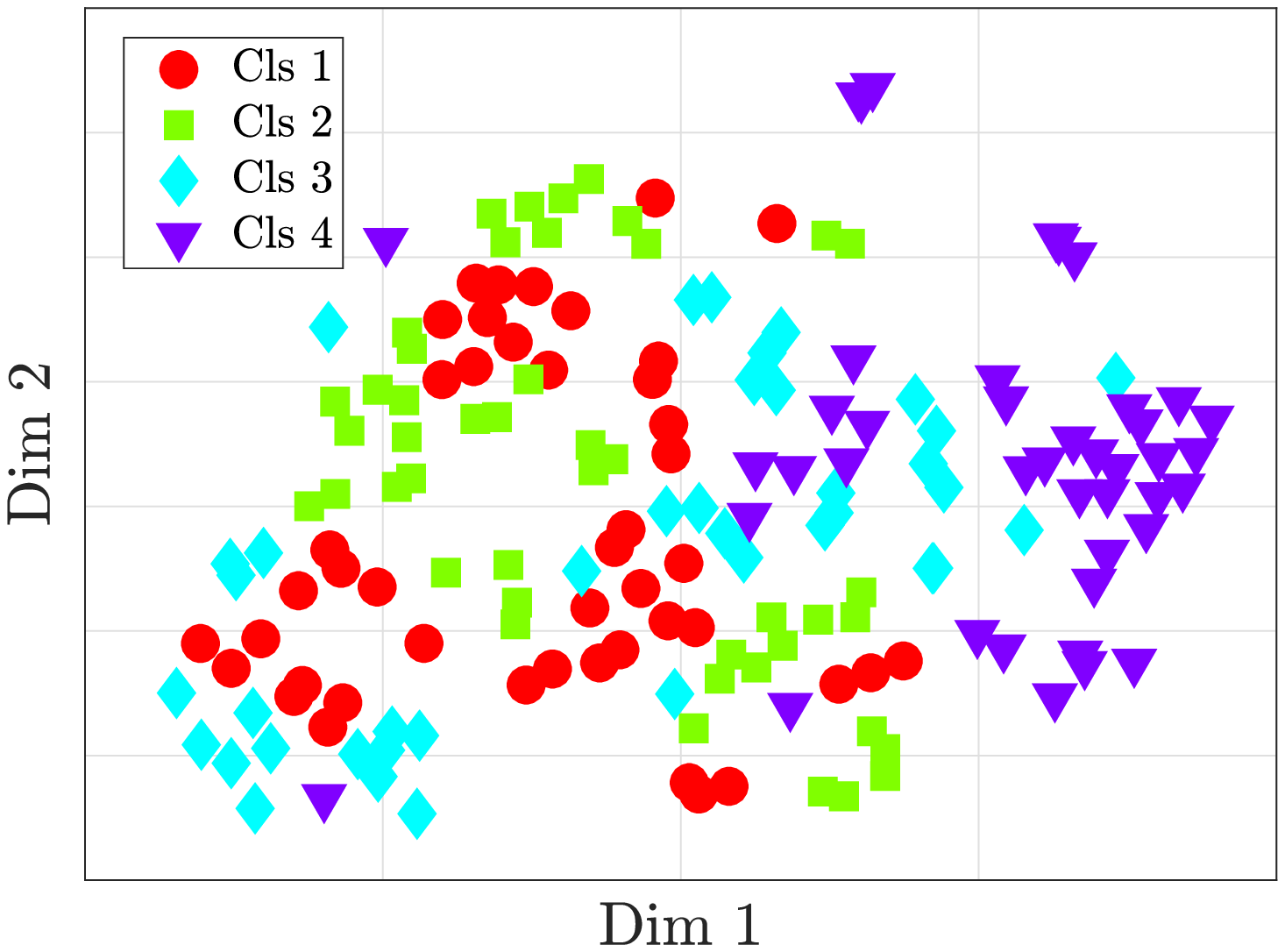}
		\endminipage
		\hspace{1cm}
		\minipage{0.45\textwidth}
		\includegraphics[width=\linewidth]{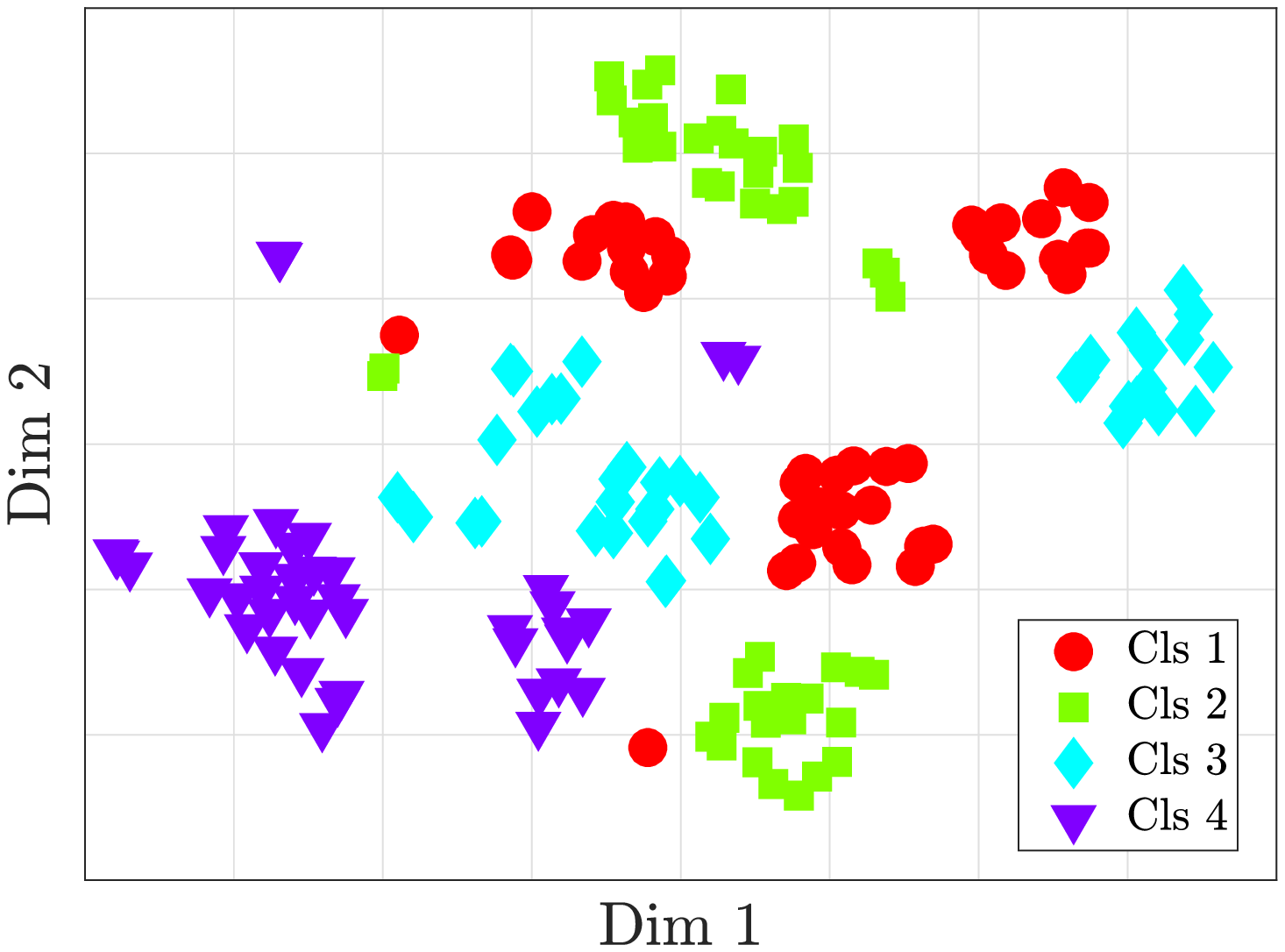}
		\endminipage
	\end{center}
	\caption{Visualization of class overlapping for the \texttt{TOX\_171} dataset based on the average-kernel combination (left) and the optimized $\nAlf$-combined embedding (right). Clearly, the kernel weighting scheme has reduced the overlapping between the classes.}
	\label{fig:tsne}
\end{figure}

To visualize the effect of the learned kernel weights ($\nAlf$) on the distribution of classes, we visualized the 2-dimensional embeddings of the \verb|TOX_171| dataset in Figure~\ref{fig:tsne} (using the t-SNE method). 
Clearly, the optimized $\nAlf$ has lead to better local separation of the classes in the resulted RKHS (Figure~\ref{fig:tsne}-left) compared to the average-kernel representation of the data ($\nAlf=\vec{1}/\nf$) in Figure~\ref{fig:tsne}-right.
This observation complies with the role of $\MC{J}_{ls}$ in Eq.~(\ref{eq:dksrc}).
\subsection{Effect of Parameter Settings}
%
%
%
%
%
We study the effect of parameters $\{\lambda, \mu, \tau, \nT \}$ 
on the $Acc$ and $Ip$ performance of IMKPL
by conducting 4 individual
experiments on the \verb|Isolet| dataset.
Each time, we change one parameter while fixing others by the values in Table~\ref{tab:data}. 

\begin{figure*}[b!]
	\begin{subfigure}{0.3\textwidth}
		\centering		
		\includegraphics[width=1\linewidth]{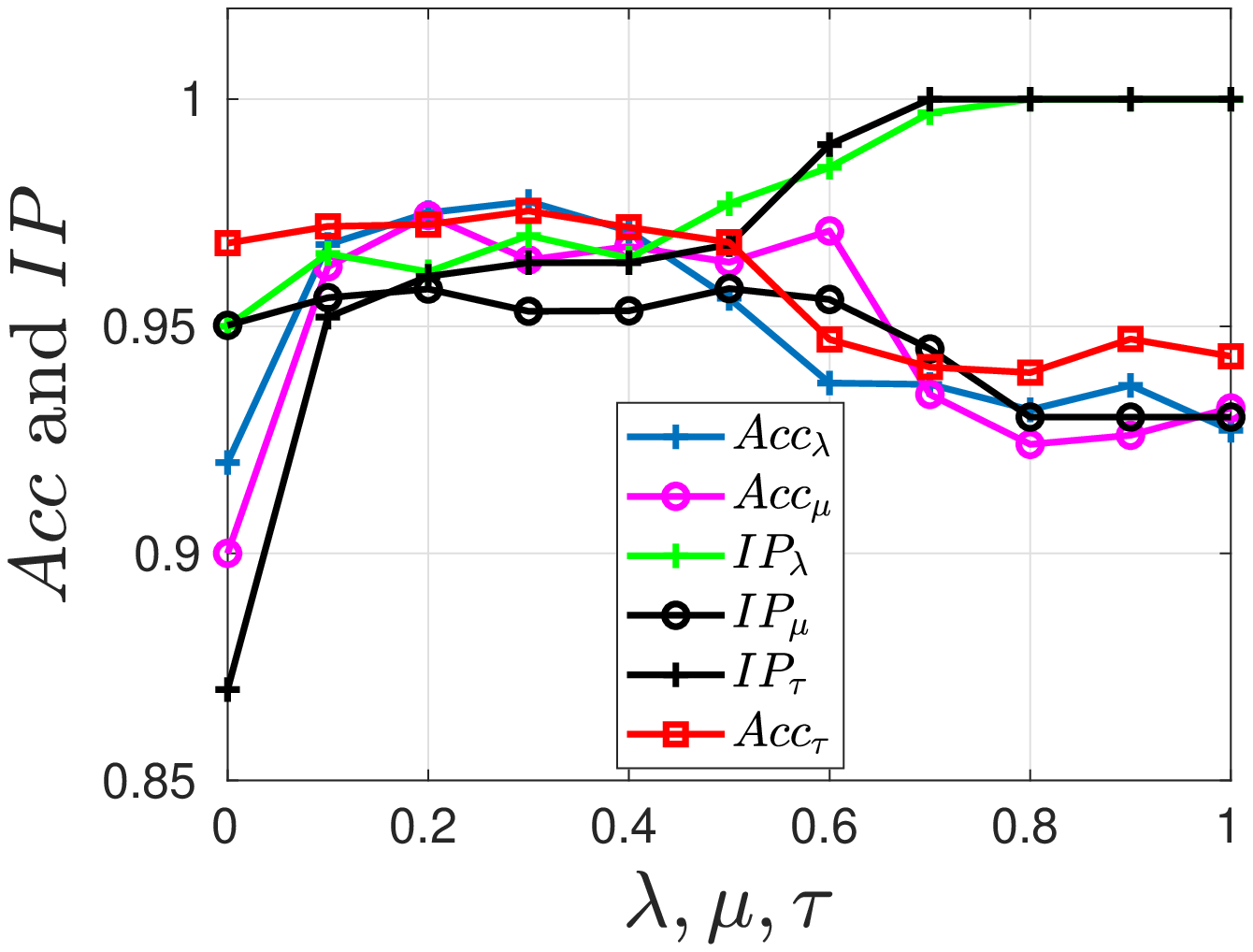}
	\end{subfigure}	
	\hfill
	\begin{subfigure}{0.3\textwidth}
		\centering		
		\includegraphics[width=1\linewidth]{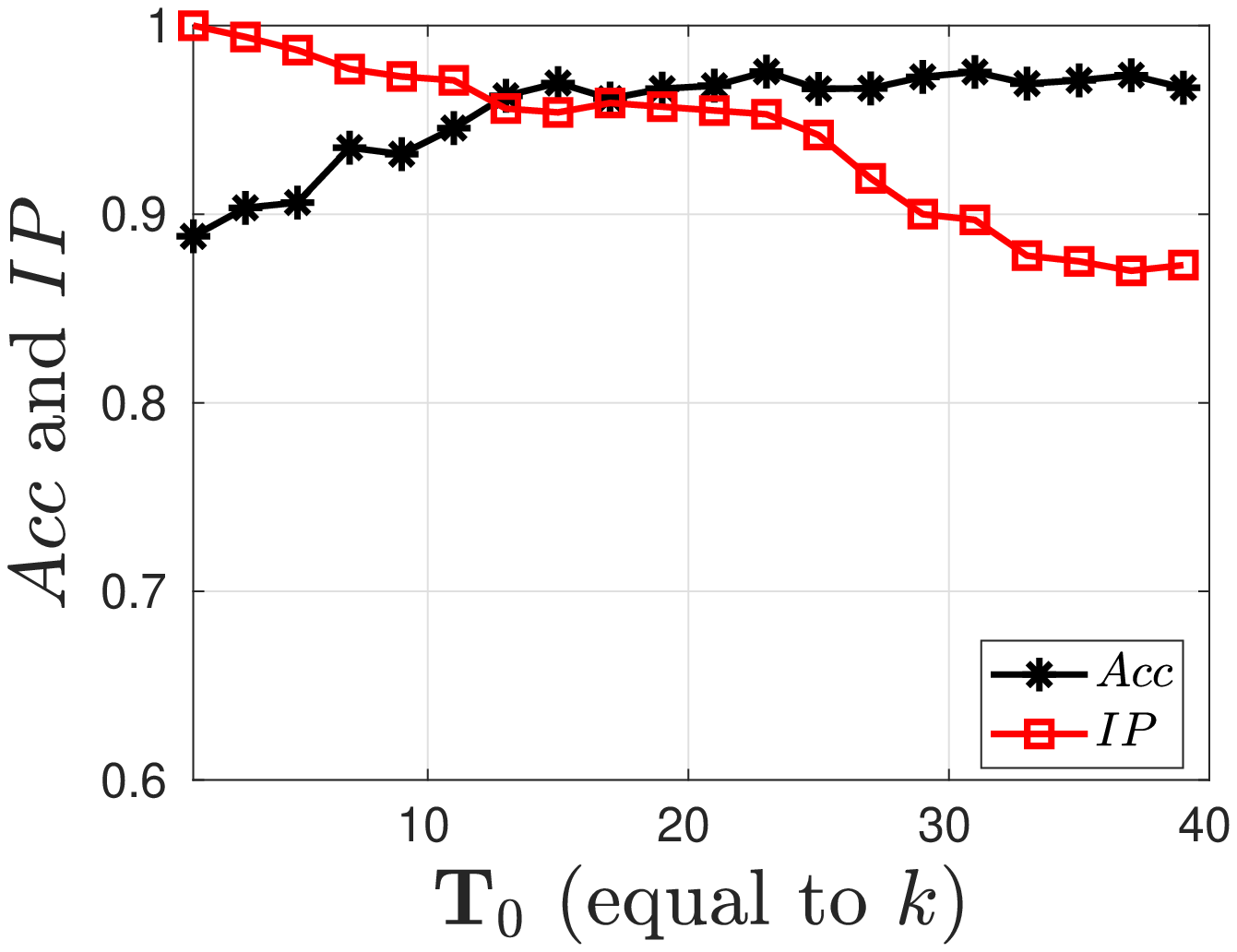}
	\end{subfigure}	
	\hfill
	\begin{subfigure}{0.3\textwidth}
		\centering		
		\includegraphics[width=1\linewidth]{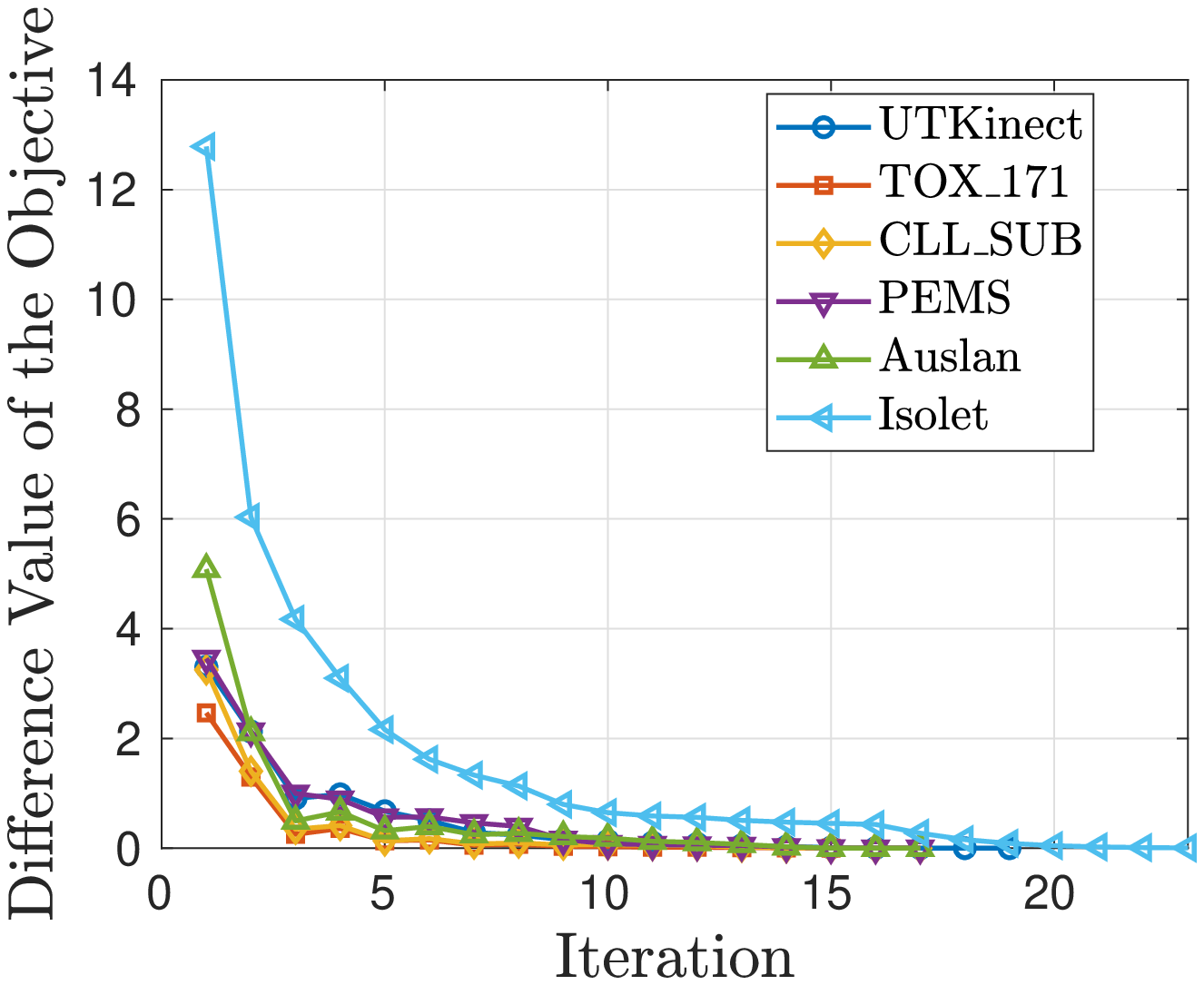}
	\end{subfigure}	
		\vspace{-0.2cm}
	\caption{Isolated effects of changing the parameters $\{\lambda,\mu,\nT,\tau\}$ on the performance measures $Acc$ and $IP$ 
		for the \texttt{Isolet} dataset, and convergence curves of IMKPL on the selected datasets.}
	\label{fig:sens}
		\vspace{-0.2cm}
\end{figure*}

As illustrated by Figure~\ref{fig:sens}-(left), the performance is acceptable when $\lambda, \mu, \tau \in [0.1~0.5]$,
but $Acc$ and $IP$ may decrease outside of this range.
Specifically, $\tau$ has a slight effect on $Acc$, but it increases the value of $IP$ almost monotonically.
In comparison, $\mu$ and $\lambda$ influence $Acc$ more significantly.
Nevertheless, they have small effects on $IP$ when they are small (in $[0~0.6]$), 
but for large values, 
$\lambda$ has a productive and $\mu$ a slight destructive effect. 
%
When the classes have considerable overlapping in the RKHS, focusing only on $\MC{J}_{ls}$ (large $\mu$) does not necessarily provide the best prototype-based solution. 
%
%
%

Figure~\ref{fig:sens}-(right) shows that increasing $\nT$ generally improves $Acc$ up to 
an upper limit.
%
Since $\ndic=\ncl\nT$, 
large values of $\nT$ 
leads to learning redundant prototypes.
%
Besides, increasing $\nT$ generally degrades the $IP$ value, 
%
but it
almost reaches 
a lower bound value for large $\nT$  ($\approx 87\%$ for \verb|Isolet|) because of the minimum interpretability induced by the non-negativity constraint $\nas_{ji}\in\MBB{R}_+$ in Eq.~(\ref{eq:dksrc}).
%
%
\subsection{Run-time and Convergence Curve}\label{sec:conv}
To evaluate the computational complexity of IMKPL, we compare the training run-time of selection methods on 
\texttt{CLL\_SUB}, \verb|AUSLAN|, and \verb|rcv1| datasets.
As reported in Table~\ref{tab:run}, \method has smaller computational time than other MK algorithms (MKLDPL, DKMLD, and MIDL) and 
is even faster than AKGLVQ (as a single-kernel method) when the number of features $\nf$ is small in relation to $N$ (\verb|AUSLAN| and \verb|rcv1|).
Although the PS algorithm has smaller run-time than IMKPL, it is not applicable to the multiple-kernel data.

In Figure~\ref{fig:sens}, we plot the difference value of the whole objective function in Eq.~(\ref{eq:dksrc}) as the difference between the objective value in each iteration and its value in the previous iteration.
Based on this figure, Algorithm~\ref{alg:mksrc} is considered converged when the above value becomes relatively small, which occurs rapidly on all the selected datasets in the experiments (less than 20 iterations). 
\section{Conclusion}\label{sec:conc}
We proposed a prototype-based learning framework to obtain a discriminative representation of datasets in the feature space.
Following
our explicit research objectives, 
IMKPL learns interpretable prototypes
as the local representatives of the classes
(e.g., a subset of similar \textit{walking} samples)
while discriminating 
the classes from each other.
%
Additionally, IMKPL performs a discriminative feature selection by finding an efficient combined embedding in feature space.
Experiments on large-scale and high-dimensional real-world benchmarks in 
both vectorial and time-series domains
validate the superiority of IMKPL over other prototype-based baselines regarding the above concerns.
\begin{table}[t]
	\centering
	\caption{Training run-time of baseline algorithms (seconds).}	
	\footnotesize
	\begin{tabularx}{0.8\textwidth}{Xcccccc} %
		\toprule	
		Dataset & \textbf{\method\tiny(ours)} & MKLDPL & DKMLD & MIDL& AKGLVQ & PS \\
		\midrule
		\texttt{CLL\_SUB}&2.58e2&2.85e4&4.08e4&8.76e4&1.24e2&2.32e0\\
		
		\verb|AUSLAN| &1.46e4&3.68e6&4.96e6&1.38e7&2.87e4&1.15e1\\
		
		\verb|rcv1|(sub1) &1.63e5&3.75e8&6.97e8&2.09e9&3.97e5&6.62e1\\
		\bottomrule		
	\end{tabularx} 				
	\label{tab:run} 
\end{table}
\newpage
\section*{Acknowledgement}
This research was supported by the Center of Cognitive 
Interaction Technology 'CITEC' (EXC 277) at Bielefeld University, which
is funded by the German Research Foundation (DFG).
\bibliographystyle{unsrt}
\bibliography{c:/Thesis/Publications/Ref4Papers_CS}



\end{document}